\documentclass{article}

\usepackage[preprint,nonatbib]{neurips_2024}

\usepackage[utf8]{inputenc} 
\usepackage[T1]{fontenc}    
\usepackage{hyperref}       
\usepackage{url}            
\usepackage{booktabs}       
\usepackage{amsfonts}       
\usepackage{nicefrac}       
\usepackage{microtype}      
\usepackage{xcolor}         
\usepackage[noline,noend, boxruled]{algorithm2e}
\usepackage{amsmath}
\usepackage{amssymb}
\usepackage{caption}
\usepackage{subcaption}
\usepackage{graphicx}
\usepackage[numbers]{natbib}
\usepackage{siunitx}
\usepackage{pifont}

\DeclareRobustCommand{\mitro}[1]{{\color{red} (Dan: #1)}}

\newcommand{\Lex}{\textsc{Lex}}
\newcommand{\Phon}{\textsc{Phon}}
\newcommand{\Context}{\textsc{Context}}
\newcommand{\Visual}{\textsc{Visual}}
\newcommand{\Motor}{\textsc{Motor}}
\newcommand{\Mood}{\textsc{Mood}}
\newcommand{\Scene}{\textsc{Scene}}

\newcommand{\TPJ}{\textsc{Role}}
\newcommand{\TPJscene}{\TPJ_\text{scene}}
\newcommand{\TPJagent}{\TPJ_\text{agent}}
\newcommand{\TPJpatient}{\TPJ_\text{patient}}
\newcommand{\TPJaction}{\TPJ_\text{action}}
\newcommand{\Subj}{\textsc{Subj}}
\newcommand{\Obj}{\textsc{Obj}}
\newcommand{\Verb}{\textsc{Verb}}

\begin{document}

\def\NEMO{{\sc Nemo}}

\title{Simulated Language Acquisition \\
in a Biologically Realistic Model of the Brain}

\author{%
  Daniel Mitropolsky \\
  McGovern Institute for Brain Research \\
  MIT \\
  Cambridge, MA \\
  \texttt{mitropol@mit.edu} \\
  \And
  Christos H. Papadimitriou \\
  Department of Computer Science \\
  Columbia University\\
  New York, NY \\
  \texttt{christos@columbia.edu} \\
}

\maketitle

\setcounter{page}{1}

\begin{abstract}
Despite tremendous progress in neuroscience, we do not have a compelling narrative for the precise way whereby the spiking of neurons in our brain results in high-level cognitive phenomena such as planning and language.  We introduce a simple mathematical formulation of six basic and broadly accepted principles of neuroscience: excitatory neurons, brain areas, random synapses, Hebbian plasticity, local inhibition, and inter-area inhibition. 
We implement a simulated neuromorphic system based on this formalism, which is capable of basic language acquisition: Starting from a tabula rasa, the system learns, in any language, the semantics of words, their syntactic role (verb versus noun), and the word order of the language, including the ability to generate novel sentences, through the exposure to a modest number of grounded sentences in the same language.  We discuss several  possible extensions and implications of this result.

\end{abstract}
\section{Introduction}

It is beyond doubt that cognitive phenomena such as language, reasoning, and planning are the direct product of the activity of neurons and synapses, and yet there is no existing overarching theory that explains exactly how this could be done. 
In the words of Richard Axel \cite{AxelNeuron2018},  {\em ''we do not have a logic for the transformation of neural activity into thought and action. I view discerning [this] logic as the most important future direction of neuroscience.''}  Through the use of the term ``logic'' in this statement, Axel seems to suggest that the research community should strive to identify, from among the ever growing volume of neuroscientific knowledge, see e.g.~\cite{Kandell}, a conceptual core of basic elements and principles which suffice for bringing about the brain’s function.  

We introduce such a minimalistic neural model of the brain, which we call \NEMO\ (see Figure 1, and \cite{PNAS}), a simple mathematical formulation of six basic and uncontroversial ideas in neuroscience: excitatory neurons, brain areas, random synaptic connectivity, local inhibition in each area, Hebbian plasticity, and interarea inhibition.  Importantly, \NEMO\ can be simulated very efficiently at the scale of tens of millions of neurons and trillions of synapses.   To test whether \NEMO\ qualifies as the abstraction sought in Axel's statement, we design a \NEMO\ system that can carry out what is arguably the most prominent achievement of the animal brain, namely {\em natural language acquisition.}  
In past work on \NEMO\ it has been shown, through both mathematical proofs and simulations, that it is capable of carrying out basic computations through the formation and manipulation of neural assemblies \cite{PNAS}, as well as various tasks of a cognitive nature \cite{DabagiaClassifier,Planning}, including {\em parsing} natural language sentences \cite{parser, CenterEmbedding}. 
However, that \NEMO\ parser presupposed that the words in the language are already represented as assemblies of neurons whose synaptic connections encode the grammatical role of each word. 

Here, we demonstrate that lexicon, syntax and semantics of a language can be {\em learned} by a tabula rasa \NEMO\ system in a way that mirrors human language acquisition. 
In particular, our system learns
 (a) semantic representations of concrete nouns and verbs; (b) basic syntactic characteristics such as the part of speech for each word; (c) the language's word order, in several moods, and (d) to generate novel sentences --- all through exposure to grounded language.

Language acquisition, of course, entails much more than this. In particular, a complete account must include the acquisition of speech sounds and word phonetics, parts-of-speech other than nouns and verbs, including functional categories of words such as ``the,'' abstract words, and syntax beyond basic word-order. Furthermore, in the baby brain the learning of all these aspects is intertwined in subtle ways \cite{lust}. 
However, since our purpose here is to design a rigorous experiment that establishes that basic language acquisition can be carried out by a \NEMO\  artifact, we focus on implementing the core statistical learning tasks (a)--(d) above 
in two phases: first the learning of word semantics and part of speech (a--b) and then the learning of word order and generation (c--d).  As for speech sounds and phonetics, we bypass this phase by adopting an input-output convention whereby a sentence is presented to the system as a sequence of stimuli corresponding to word tokens, and similarly for sentence generation.

\begin{figure} \label{figure:nemo-model}
    \centering
    \includegraphics[width=0.9\linewidth]{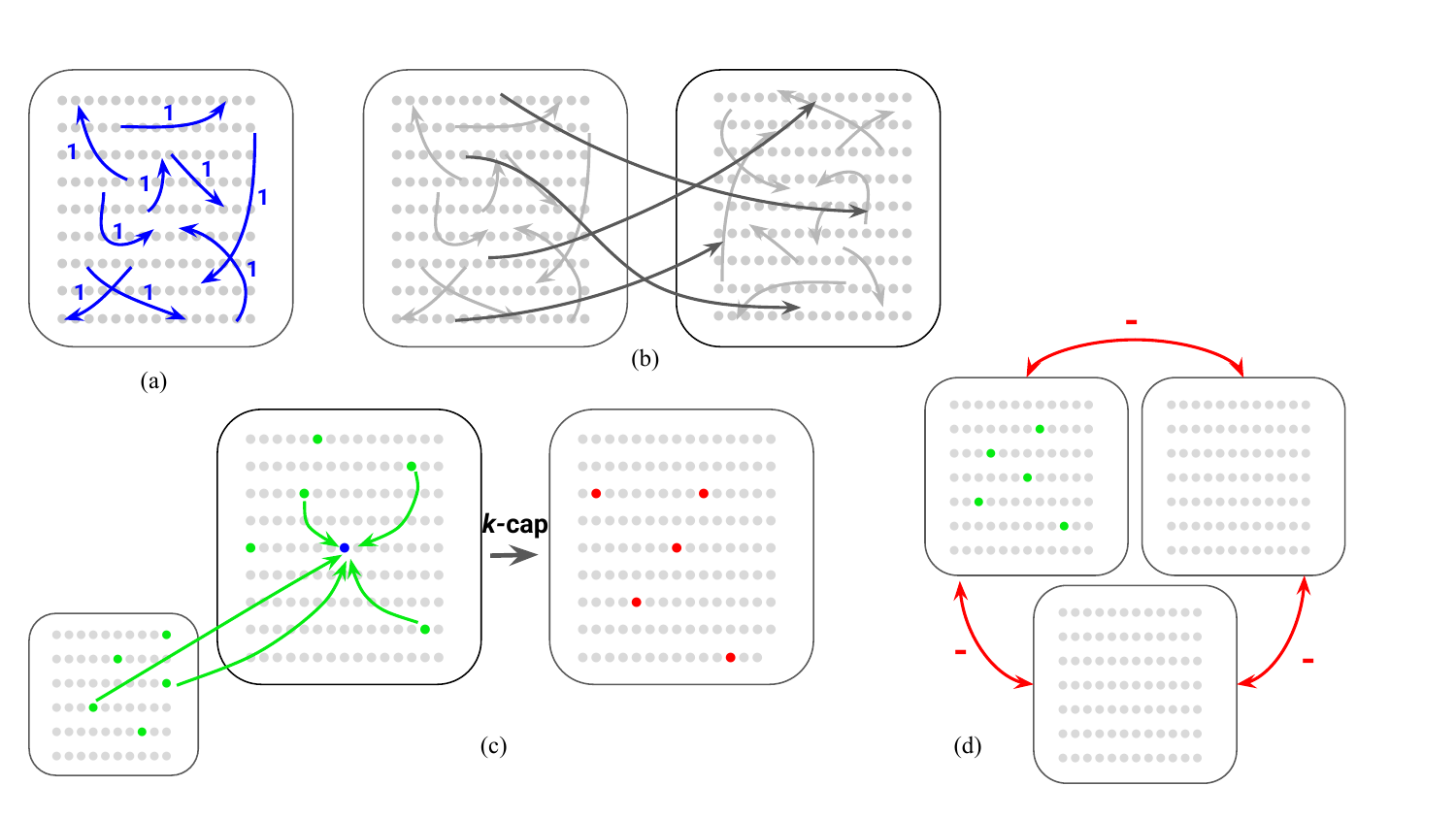}
\caption{An overview of our biologically constrained neural model (\NEMO\ for short).  The brain is modeled as a set of brain \emph{areas}, each of which (a) has $n$ neurons; in our experiments $n$ is typically a million. For every ordered pair $(i,j)$ of neurons in an area, a directed synapse from $i$ to $j$ is included with probability $p$, initialized with weight $1$, resulting in a directed random synaptic graph; in our experiments, $p$ is typically equal to $0.01$. (b) \NEMO\ artifacts consist of multiple areas; each ordered pair of areas may be connected by a {\em fiber} in which case each possible synapse from each neuron in the first area to each neuron in the second is included with probability $p$.  Only a few of all possible pairs of brain areas are connected by a fiber, defining the top level of the system's architecture. (c) \NEMO\ is a dynamical system defined by the \emph{firing} operation. Time proceeds in discrete steps. At each time step only $k$ neurons in each area fire --- a selection that models local inhibition in the area --- where $k$ is much smaller than $n$ (in our experiments, $k$ is typically about $\sqrt{n}$). Of all $n$ neurons in an area the neurons that fire are those $k$ that have received the {\em largest synaptic input} from the previous time-step.  That is, for each neuron $i$ the quantity $\sum_{\text{synapses}~(j,i)} w_{j,i}$ is computed, where the sum is taken over all neurons $j$, in the same area or in other areas, that fired in the previous step and have a synapse to $i$, and $w_{j,i}$ stands for the weight of this synapse. Once these sums have been computed for all neurons in the area, the top $k$ neurons are selected to fire in the next time step, with any ties resolved arbitrarily. This primitive of the \NEMO\ model is called $k$-{\em cap}. In the illustration above, to the left of the ``$k$-cap" arrow we show an area with firing neurons (green), as well as firing neurons in another area, and an example neuron (dark blue) showing its incoming edges from firing neurons (such edges from firing neurons to other neurons are not shown). To the right of the ``$k$-cap" arrow is the state of the area after inputs have been summed and ``$k$-cap" applied; the dark blue example neuron from the previous time step happened to be among the five top-$k$ inputs, and is hence firing (red); in this example, $k=5$. There is also  {\em Hebbian plasticity:} for two synaptically connected neurons $j,i$ firing in succession for two consecutive steps in this order, the synaptic weight $w_{j,i}$ is multiplied by $(1+\beta)$; the positive plasticity parameter $\beta$ is typically $0.05$. (d) Additionally, two or more areas can be in \emph{mutual inhibition}, in which case there is firing only in the area that receives the greatest total synaptic input from the previous step, and no firing in the other areas. The set of areas, fibers, mutually-inhibited sets of areas, as well as $n,p,k,$ and $\beta$ are the \emph{hyperparameters} of the \NEMO\ model.  In our experiments, a local hyperparameter such as $p$ and $\beta$ may vary from one area or fiber to the other.}
    \label{fig:O-core}
\end{figure}

In our experiments, the learning device is exposed to {\em grounded whole sentences} in any fixed natural language (we use several languages in our experiments).  By ``grounded'' we mean that each sentence is presented while representations corresponding to the sentence are active in the sensory and motor cortex.  That is, when the sentence ``the dog runs'' is input, we assume that in the premotor cortex the mirror cells \cite{KemmererGonzalez2010, Fernandino2010} for ``run,''  as well as the generic dog representation in the inferotemporal visual cortex \cite{KarDiCarlo}, are firing. 

In a very interesting precursor to this work, it was shown in \cite{TomaselloPulverMueller} that concrete nouns, verbs and their meanings can be learned through grounded presentation in a different biologically realistic model of the brain.  However, that work was about the easier problem of learning {\em isolated words,} without syntactic context --- it is known that nonhuman primates can achieve this level of language learning, see, for example \cite{SeidenbergApesSigning}.  Here we solve the far more demanding problem --- and an exclusive characteristic of humans --- of learning word semantics and the verb vs. noun distinction, as well as word order, through exposure to grounded whole sentences. We believe that our results entail the first biologically plausible artifact that is capable of basic language acquisition; if scientists were to discover such linguistic competence in a nonhuman animal, we believe that this finding would fundamentally change the currently held view on the exclusivity of language.

Besides addressing the quest for Axel's ``logic'' mentioned above, the present work also pertains to another long-standing fundamental question, namely: {\em What is the biological basis of language?} In particular, is language the fruit of unique human characteristics at the genetic, molecular, and/or neural level?  In the past, the discovery of such characteristics (e.g., the FOXP2 gene, for a brief time called ``the language gene,'' and the novel dendritic potentials recently discovered in human layer 2-3 neurons \cite{gidon2020}) has led to speculation about the biological roots of the human linguistic ability, see, e.g., \cite{Dehaene2022}. The biologically plausible language acquisition system presented here can be seen as evidence for an alternative {\em null hypothesis:} A language system { can} be built on top of the mammalian brain, through simple and widely accepted neuroscientific elements and principles.

   

The specific neural architecture and algorithms we propose can be seen as a comprehensive, neurobiologically plausible {\em hypothesis} for the human language system, a domain in which such concrete hypotheses are scarce. The brain areas used in our model --- their number, function and synaptic connections between them --- are arguably compatible with existing neuroanatomical and psycholinguistic evidence and opinion, and reflect the field's consensus where it exists.  In fact,  our artificial language system can be the source of predictions about activations in these brain areas in the course of comprehension and generation, which may prove useful in neurolinguistic research.  

Finally, let us note here that, compared to another extant class of computational artifacts with linguistic abilities, namely the large language models (LLMs) of now \cite{LLMoverview}, our system has the important distinction of {\em biological plausibility} --- in particular, it does not use backpropagation, the most important ingredient of current machine learning, which has never been observed in the animal brain.

\bigskip

\section{Procedures and Results}
\subsection{Learning word semantics}
We start by describing how the \NEMO\ system for language acquisition learns the meaning of concrete nouns and intransitive verbs, as well as their syntactic role (noun versus verb), see Figure \ref{fig:semantic-learning}, and see SI for the parameter values.  In the next section we describe the extensions of this core system that are needed for learning syntax and the word order of the language (compare Figures 2 and 4).  

The earliest part of language acquisition during roughly the first year of life involves learning the speech sounds of the language.  In our experiment we bypass this stage by adopting a convenient {\em input/output convention:} Words are input to the device through the activation of word representations in an area that we call $\Phon$, where phonetic knowledge for recognizing and articulating words is already in place.  There seems to be a consensus in neurolinguistic literature (see, e.g., \cite{Hicock2003, OKADA2006112}) that the human brain indeed contains such an area. $\Phon$ an {\em input area}, in the sense that it is initialized with assemblies, each representing the phonetics of a word in the language. This input-output convention is tantamount to assuming that the device knows {\em ab ovo} how to recognize and articulate the words in the lexicon, but at the start it is oblivious of their meaning or part of speech.

The {\em lexicon} consists of two areas, $\Lex_1$ and $\Lex_2$, which start as a tabula rasa, but in the experiment will end up forming one assembly representation for each word in the lexicon: in $\Lex_1$ for nouns and in $\Lex_2$ for verbs. This reflects a hypothesis in neurolonguistics, based on neurological studies and separation studies between nouns and verbs, that nouns and verbs are represented in different areas of the brain \cite{Matzig2009, Vigliocco2011, Kemmerer2012}. There are fibers that go, both ways, between $\Phon$ and each of $\Lex_1$ and $\Lex_2$. 

The acquisition system also has a number of \emph{semantic areas}, sensorimotor and other cortical areas that feed into the language system; see, e.g., \cite{GallantSemantics} for results suggesting the existence of many such areas. Two of these areas, $\Visual$ and $\Motor$, have special significance, but there are also several additional context areas representing other modes of semantic context such as olfactory, non-linguistic auditory, location, emotion, etc., and labeled $C_i$ for $i=1,\ldots,m$.  Like $\Phon$, the semantic areas are input areas, initialized to contain assembly representations that provide semantic content to words in the lexicon, with action verbs taking their meaning predominantly from mirror neurons in the motor cortex and concrete nouns from visual representations in the visual cortex. The exact total number $m$ of extra context areas will vary in our experiment; our algorithm works for various $m$, from zero to ten.

There are fibers connecting the lexical and semantic areas --- it is known that there are many evolutionarily novel fibers in the vicinity of the left medial temporal lobe where the lexicon is believed to reside in the human brain, see e.g.~\cite{friedericiPathways}.  In our model there are fibers in both directions between $\Lex_1$ and $\Lex_2$ on the one side, and $\Phon$ and the semantic areas on the other.  Four of these $2m+6$ fibers, namely the ones between $\Phon$ and the two lexical areas, as well as the one between $\Lex_1$ and $\Visual$, and the one between $\Lex_2$ and $\Motor$, have increased parameters $\beta$ and $p$, making them stronger conduits of synaptic input --- these fibers are depicted as bolder arrows in the figure. 


\begin{figure} \label{fig:semantic-learning}
    \centering
    \includegraphics[trim={8cm 5cm 8cm 5cm},clip,width=0.5\linewidth]{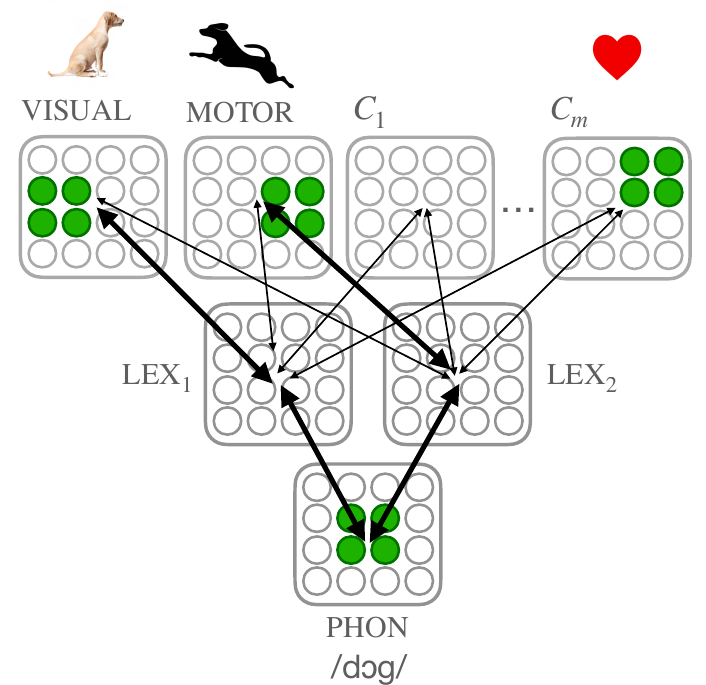}
    \caption{The \NEMO\ system for learning concrete nouns and verbs. $\Phon$ is an input area which is initialized with one  representation, for each word, of the phonological code of the word.  The semantic areas $\Visual$, $\Motor$ and $C_1,\ldots,C_m$ are also input areas, with  visual representations, motor representations, and other perceptual representations (e.g. olfactory, emotional, etc). $\Lex_1$ and $\Lex_2$ are the areas where the nouns and verbs, respectively, in the language will be learned.  $\Lex_1$ has high connectivity to $\Visual$ and $\Lex_2$ to $\Motor$, and lower connectivity to the other semantic areas. The figure shows the firing pattern in the input areas when input the sentence ``dog jumps,'' at the moment ``dog'' is heard. }
    \label{fig:semantic-learning}
\end{figure}


\paragraph{The word learning experiment.} \label{subsection:word-learning}
We fix a natural language and $\ell$ concrete nouns and $\ell$ intransitive action verbs, where $\ell$ is a parameter that we vary in our experiment. By restricting ourselves to action nouns and concrete verbs we assure that for each word in the lexicon there is a corresponding semantic representation in the semantic areas. The restriction to intransitive verbs is considered only for the present experiment of learning word semantics; transitive verbs will be introduced soon.  All sentences in the experiment are of length two: ``the cat jumps'' and ``the dog eats,'' where, for the purposes of our experiment, we assume that function words such as ``the'' are simply ignored; taking into account such words makes the learning task easier
. Note that the language can have either SV (subject-verb) principal word order (as in English, Chinese and Swahili) or VS (as in Irish, Modern Standard Arabic, and Tagalog), and our model should succeed in either scenario; this is one of the main challenges of this experiment.  

For each word $w$ we initialize an assembly $\Phon[w]$ in $\Phon$.  In addition, if $w$ is a noun, we initialize an assembly in $\Visual$ corresponding to the visual representation of the object of $w$ in the inferotemporal cortex,  and if $w$ is a verb, there is a corresponding assembly in $\Motor$ --- presumably the mirror cells for $w$ in the premotor cortex.  Finally, for a subset of the remaining $m$ context areas, selected at random, we create in each area an assembly corresponding to $w$. 

We input randomly generated sentences via a uniform choice of noun and verb.  Note that this results in including bizarre sentences and impossible scenes, such as ``the trees jump'', but we have verified that omitting these does not affect the results. To input a sentence, we input each word individually, in the word order of the language, by firing the corresponding assembly in $\Phon$ $\tau$ times, where $\tau$ is a parameter of the experiment, while the contextual input of the scene (the assemblies associated with the words in the sentence, both with the noun and the verb) fire throughout the duration of the entire sentence. That is, as in life, for the sentence ``the cat jumps,'' the $\Visual$ assembly representing ``cat'' and the $\Motor$ assembly representing ``jump'' both fire as the learner hears ``cat'' and then ``jumps.''  Note that it is a priori impossible to know which of the two words is the concrete noun, and which is the action verb. However, as we shall see, the interaction between \NEMO\ dynamics and cooccurrence statistics results in the system learning to correctly distinguish between noun and verb.  

What does it mean for the system to succeed in learning these nouns and verbs? 
Informally, we would like $\Lex_1$ to become a ``noun area,'' and $\Lex_2$ a ``verb area,'' containing \emph{robust neural representations of each noun and verb} in the lexicon, respectively, and we would also like each of these robust word representations to be somehow aware of the meaning of the corresponding word, as well as aware of their affinity with the corresponding phonetic assembly in $\Phon$.  

More formally, we require that, after the input of enough sentences, the following properties hold:

\paragraph{Property 1: Formation of word assemblies.} For each noun, if we fire its representation in $\Phon$, or that in $\Visual$, or both, the resulting $k$-cap in $\Lex_1$ is so densely interconnected that, if we fire it by itself, the $k$-cap in $\Lex_1$ remains the same set of neurons again and again; and that this also occurs if we continue to fire the representation in $\Visual$. In the same sense, a neural assembly emerges for each verb in $\Lex_2$. 

\paragraph{Property 2: Formation of the PHON-LEX-VISUAL pathway.} The creation of assemblies for each word enables a pathway between $\Phon$ and the semantic areas via $\Lex_1$ and $\Lex_2$. The pathway is bidirectional: for each noun $w$, if we fire $\Phon[w]$ for $2$ time steps, via $\Lex_1$ we activate $\Visual[w]$ and its representation in the contextual areas. Conversely, if we fire $\Visual[w]$ and the contextual representations, we retrieve $\Phon[w]$. In the same way, a pathway is forged for verbs.

\paragraph{Property 3: Classification of Nouns vs. Verbs} The final property required to emerge from learning is that, while very robust assemblies for nouns are formed in $\Lex_1$ and for verbs in $\Lex_2$, \emph{nothing akin} to a stable assembly representation exists for nouns in $\Lex_2$, nor for verbs in $\Lex_1$. For a noun $w$, firing $\Phon[w]$ into $\Lex_2$ results in a ``wobbly'' set of neurons: firing this set again recurrently, \emph{even with} the input of $\Phon[w]$, results in a quite different set. Additionally, firing this set \emph{back} into $\Phon$ results in a set that has very little overlap with $\Phon[w]$. This last property yields an algorithmic test for whether a word is a noun or verb: try firing $\Phon[w]$ into $\Lex_1$ and $\Lex_2$, and see which of the two $k$-caps activates $\Phon[w]$ when firing back into $\Phon$.




\subsection{Results of the word learning experiment}
We experiment with different word orders and lexicon sizes, and we input to the system two-word sentences, as described above, {\em until all three Properties 1-3 above hold.}  
We are interested in determining how many sentences are needed for this, as a function of the number $\ell$ of nouns and verbs in the lexicon.  Recall that the amount of linguistic input required for language to be learned is of central interest in linguistics, for example, in relation to the ``poverty of stimulus'' argument \cite{Chomsky1986}. That is, our experiment offers an opportunity to actually quantify what ``modest linguistic input'' means in the context of language learning. We propose that one useful definition of modest linguistic input is $O(\ell)$ sentences, where $\ell$ is the size of the lexicon --- obviously at least that many sentences are needed. 

\begin{figure}
    \centering
    \begin{subfigure}[b]{0.49\textwidth}
         \centering
         \includegraphics[trim={0.5cm 0cm 1cm 1cm},clip,width=\textwidth]{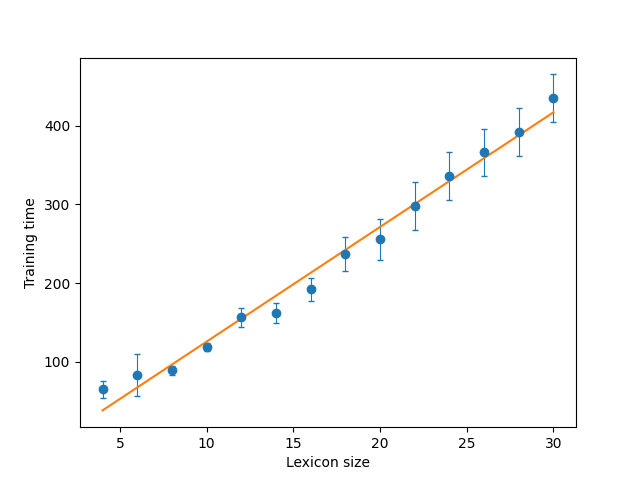}
         \caption{}
         \label{fig:lex}
    \end{subfigure}
   \hfill
    \begin{subfigure}[b]{0.49\textwidth}
         \centering
         \includegraphics[trim={0.5cm 0cm 1cm 1cm},clip,width=\textwidth]{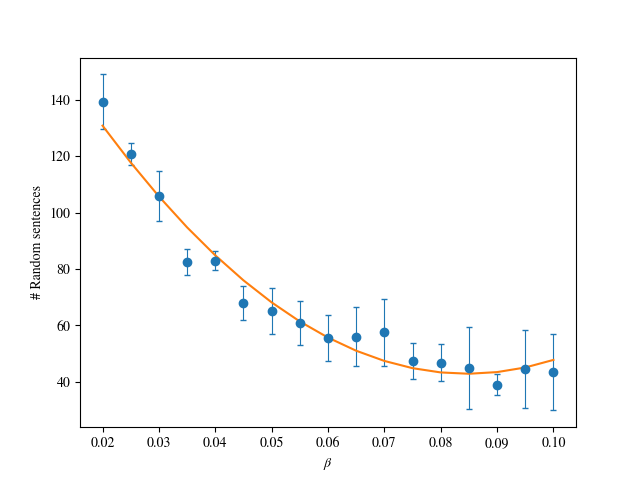}
         \caption{}
         \label{fig:betas}
    \end{subfigure}
    \\
    \begin{subfigure}[b]{0.49\textwidth}
         \centering
         \includegraphics[trim={6cm 6cm 9cm 6cm},clip,width=\textwidth]{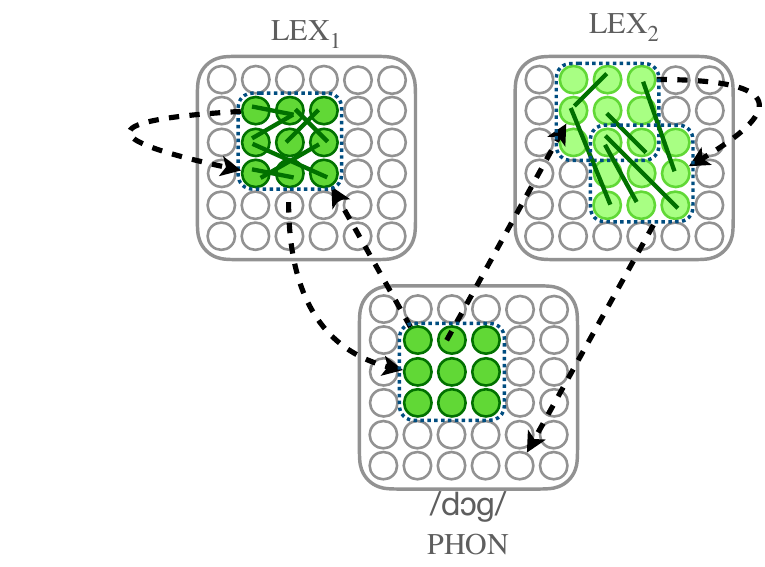}
         \caption{}
         \label{fig:lex-overlap}
    \end{subfigure}
    \begin{subfigure}[b]{0.49\textwidth}
         \centering
         \includegraphics[trim={0.5cm 0cm 1cm 1cm},clip,width=\textwidth]{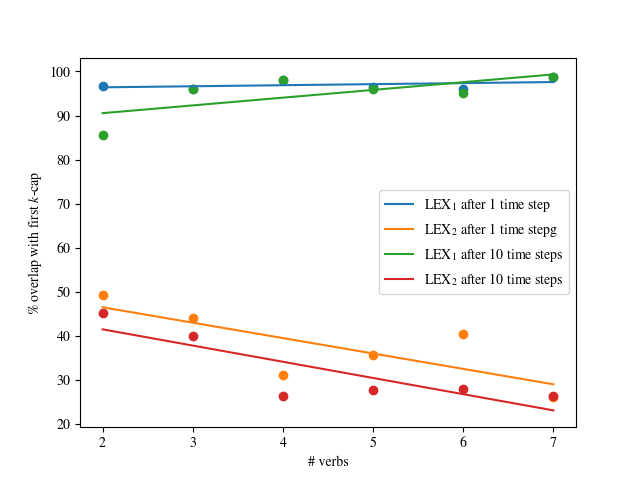}
         \caption{``dog" in $\Lex_1$ vs. $\Lex_2$}
         \label{fig:lex-overlap}
    \end{subfigure}
    \caption{Results of the word learning experiment. In (a) the learning simulation is performed for varying sizes of the lexicon until all of Properties 1-3 hold for the first time for every word, revealing a linear trend (parameters: $n=10^5,p=0.05,\beta=0.06,k_{\Lex_2} = k_{\Lex_1}=50,k_{\Context_i}=20$, and $k=100$ for other areas, and $\tau=2$). In (b) the learning experiment is repeated for varying $\beta$ always for a lexicon of size $4$. Learning is accelerated with higher plasticity until a point of saturation. (c) illustrates the formation of a stable assembly (dark green) representing ``dog'' in $\Lex_1$; in contrast, an unstable and loosely connected set of neurons (pale green) is formed in $\Lex_2$. In (d)  the same stability/instability pattern is illustrated numerically, for a varying number of verbs with which ``dog'' is paired in training sentences.  We compare the $k$-caps obtained by firing $\Phon$[dog], into $\Lex_1$ and $\Lex_2$, and then continuing to fire (both $\Phon$[dog] 
    and the $\Lex$ areas recurrently). Stability is measured in terms of the overlap of the $k$-cap obtained at the first firing of $\Phon[dog]$ and that of subsequent firings (one step and ten steps later).  The set of firing neurons in $\Lex_1$ is stable, overlapping fully with the first $k$-cap, while in $\Lex_2$ the overlap is much lower and decreases with more input verbs.}
    \label{fig}
\end{figure}

We find that, indeed, the required number of sentences appears to \emph{grow as a linear function} of the number of words in the lexicon, with a reasonable constant around ten (see Figure \ref{fig:lex}). That is, with just around $10$ sentences per word, the model (1) converges to stable, self-firing assemblies for each word in the lexicon, (2) correctly classifies nouns vs. verbs by whether an assembly is formed in $\Lex_1$ or $\Lex_2$, and \emph{not} the other, and (3) creating a pathway between $\Phon$, the lexical areas, and the semantic areas that allows for going from one representation to another (e.g. visual stimulus of a dog to $\Phon$[``dog"]). We find, perhaps unsurprisingly, that increasing the plasticity parameter speeds up learning, with a natural lower bound in the number of sentences needed to form stable and unstable representations \ref{fig:betas}. We also find that the $k$-cap of a word in the \emph{wrong} lexical area, e.g. from firing $\Phon$[``dog"] into $\Lex_2$, becomes increasingly unstable (i.e., less assembly-like) as the number of ``complements" that this word occurs with increases; for instance, if ``dog" occurs with just 7 different verbs in training, its $k$-cap into $\Lex_2$ is a set with only $30\%$ stability vs. $100\%$ in $\Lex_1$ (see \ref{fig:lex-overlap}). 

Finally, we run another variant, in which input sentences with intransitive verbs are intermingled with sentences with transitive verbs, such as ``the boy pushes the train'' containing two nouns. We find that, if the transitive sentences are not more than roughly a half of the input sentences, such a mix of linguistic input also leads to learning of semantics and syntactic roles.  In other words, even though the the first experiment is meant for intransitive sentences, the presence of some transitive sentences in the training --- which is likely in actual acquisition --- does not confound this stage of learning too much. 

\subsection{Syntax: learning the word order, and learning to generate sentences}

\begin{figure}
    \centering
    \includegraphics[scale=0.5,trim={4cm 6cm 2cm 4cm},clip]{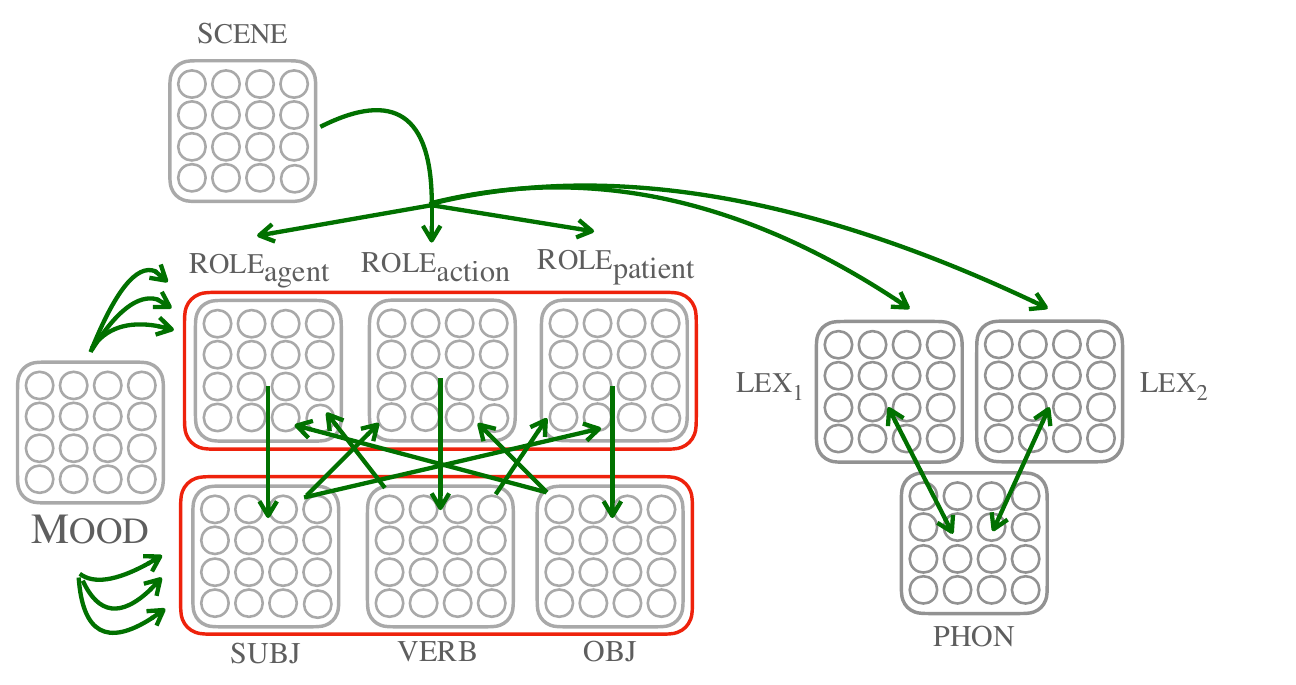}
    \caption{The \NEMO\ system for constituent order learning, as described in \ref{subsection:word-order-model}. Red boxes represent two triples of areas under mutual-inhibition. The short arrows out of $\Mood$ depict fibers into each TPJ area, as well as $\Subj, \Verb, \Obj$. The semantic areas ($\Visual$, $\Motor$, etc.) are still part of the model, but are not shown for clarity.
    }
    \label{fig:model-full}
\end{figure}
After the stage where the meaning and syntactic role (noun vs. verb) of words has been learned, we turn to the key problem of learning {\em syntax.}  Syntax encompasses an entire complex of interrelated linguistic phenomena, and of these we shall focus here on one that is particularly significant: we want our system to learn the language's basic \emph{word order}. To have learned the word order means that the system can {\em produce novel sentences} when appropriate sensorimotor stimuli are presented --- for example, in English, to articulate the sentence ``Sister pushes Dad'' upon sensory input of the corresponding scene, even if this scene was never seen or described before.

\paragraph{Word orders and grammatical moods.} Languages can be broadly classified as belonging to one of six basic categories based on the default ordering of the sentence's subject, object, and verb, known as the \emph{constituent order} or {\em word order.}  For instance, English is an SVO (subject-verb-object) language and so is French, while Japanese is SOV and Tagalog is VSO. Also of interest is the constituent ordering in intransitive sentences (SV vs. VS).  The word order is in many ways the most important syntactic characteristic of a language, and is known to determine a large part of the systematic variation in language acquisition \cite{lust}.  

We shall discuss the subject of word orders extensively, but let us note here two apparent linguistic universals \cite{wals}: (a) Word orders starting with O are extremely rare; and (b) in attested languages with a rigid or default word order, the intransitive order is almost always the same ordering with O omitted\footnote{Based off the World Atlas of Language Structures Online \cite{wals}, out of 1185 languages with default transitive and intransitive orders recorded, we found at most $11$ that \emph{may} invert $S$ and $V$ in intransitive order relative to transitive order.} --- for example, if the default transitive order is SOV, the intransitive order is SV.  As we shall see, these two linguistic facts {\em can be predicted by the workings of our device:}   our device has more difficulty learning the syntax of word orders starting with O, and of languages that violate the second property.  

We also note that many languages have varying constituent orders depending on semantic factors such as \emph{mood} --- whether the sentence is a statement, question, command, etc.~---, syntactic factors (e.g., as part of a dependent clause), or just as a result of emphasis and pragmatics (especially in languages with relatively free word order such as Polish). While we do not set out to handle the entire scope of these complexities, we will show that our model can acquire and output different word orders for different grammatical moods (declarative, interrogative, subjunctive, etc.) via a simple mechanism; we believe that this mechanism can be extended to model syntactic and pragmatic variations of the word order.

\paragraph{Thematic roles.} Analyzing a complex scene by assigning thematic roles to the participants of the scene --- determining who is doing what to whom, and who is watching --- is an important skill for animal survival. A classical study by Norman Geschwind in 1970 \cite{Geschwind1970} showed that Broca's aphasiacs easily select the correct image for a sentence like ``the lion chases the deer" and struggle to do so for ``the lion chases the tiger,'' which is consistent with a separate computation of roles in a scene based on visual cues, which helps in the former case but is confounded in the second. The precise seat of the non-linguistic mechanism that assigns roles to perceived agents in a scene is unknown, but a related kind of computation takes place in the left temporoparietal junction (lTPJ), which is believed to be a multimodal hub integrating information from diverse perceptual pathways to ``create coherent, spatiotemporal representations of the complex dynamic situations'' \cite{BoylanAngularGyrus}. The TPJ is often identified as the seat of theory-of-mind in the brain \cite{Saxe2003}: computing the knowledge states of participants and perspective taking; note that the computation of thematic roles in an observed scene is related to theory-of-mind. A recent metareview of the TPJ \cite{igelstromTPJ} concludes that the TPJ is not very involved during conscious reflection about who did what to whom; rather, it appears to carry out a lower-order and automatic processing of perceptual inputs (visual, motor, etc.) in a domain-general way, that is, as the basis for other processes in the brain, e.g. language and social cognition). It was found in \cite{Pallier2011} that the left TPJ is one of three brain areas that were implicated in the structural analysis of sentences with meaningful words, and not with jabberwocky words.  Stephen Pinker hypothesized \cite{pinker2010language} that the TPJ may be the seat of ``mentalese", the nonverbal language of thought. Humans appear to have evolved sophisticated mentalizing abilities through a proliferation in grey matter volume in the temporal cortex, and the TPJ is the most enlarged part of the temporal cortex in humans versus apes \cite{Braunsdorf2021}. It would seem that the TPJ would be most involved in mental scene ``preparation", that is, assigning the roles of agent, patient, observer, etc. to an internal mental scene (perhaps in preparation for speech). The analogous structure in macaques, called SPS, primarily codes attention to other participants, arguably a subtask of role assignment and of the theory of mind \cite{Mars2013}.

In view of the discussion above, we hypothesize that there is a system in the primate brain, not inherently linguistic and perhaps in communication with the TPJ, which computes thematic roles of scenes. Our \NEMO\ implementation contains three role areas, $\TPJagent$, $\TPJaction$, and $\TPJpatient$, 
as well as a fourth area, $\TPJscene$ synaptically connected to all three.  We assume that, when a scene is presented, such as a dog chasing a cat, the four areas are prepared as follows:  There are assemblies in the three role areas $\TPJagent$, $\TPJaction$, and $\TPJpatient$ synaptically connected to the assemblies for ``dog,'' ``chase,'' and ``cat'' in $\Lex_1$ and $\Lex_2$ (or indirectly thought $\Visual$ and $\Motor$, respectively), while $\TPJscene$ contains an assembly, representing the whole scene, synaptically connected to these three.

The question we shall pursue is this: Assuming that the \TPJ~areas prepare scene representations as described above (as we presume can be done in all primates), can the constituent order of a human language be learned from input sentences, so as to correctly generate sentences when a new scene is observed or imagined?

 \def\MOOD{\sc Mood}

\paragraph{The full language system.} \label{subsection:word-order-model}  The full system (Figure \ref{fig:model-full}) 
has several areas in addition to those of the core word learning system of Figure 2:  (a) the four {\em \TPJ\ areas:} $\TPJagent, \TPJpatient, \TPJaction$, and $\TPJscene$; (b) three \emph{syntactic} areas, $\Subj, \Obj$ and $\Verb$, and (c) a $\Mood$ area. There are $18$ fibers, shown in Figure \ref{fig:model-full}, which connect the lexical areas to and from \TPJ\ areas, the \TPJ\ areas to the syntactic areas, and $\Mood$ with the \TPJ\ areas and syntactic areas. Finally, the three \TPJ\ areas are in \emph{mutual inhibition;} this is the only use of interarea inhibition in our model.

\subsection{The syntax learning experiment}
In this section we assume the word learning experiment of Section \ref{subsection:word-learning} has previously taken place\footnote{This is done for the sake of clarity; the two experiments can be intertwined at the cost of more training, as shown in the SI.}. For learning constituent order, we input grounded sentences, both transitive and intransitive.  Grounding now means that, with each input sentence, e.g. ``Sister grabs the toy,'' a scene to the same effect has been input to the sensory system, and the thematic roles (patient, agent, action) of the scene have been computed and are represented as assemblies in the corresponding \TPJ\ areas, with strong synaptic connections to the corresponding assemblies in $\Lex_1$ and $\Lex_2$ (the latter, possibly through the context areas).
For each consecutive word in the sentence, the corresponding assembly in $\Phon$ is fired for $\tau$ time steps; this allows firing to propagate throughout the lexical areas, \TPJ\ areas, and finally into the syntactic areas. The constituent order is learned through plasticity in the synapses between the syntactic areas and the \TPJ\ role areas. For instance, if the first constituent is the subject, then the firing of the noun's assembly in $\Phon$ propagates through $\Lex_1$, to $\TPJagent$, and finally into $\Subj$. This chain of assemblies continues to fire until the next word, say the verb, is input. Again starting from $\Phon$, the firing of the verb's phonological assembly will propagate through firing into $\TPJaction$; at this point, neurons from $\Subj$ (from the previous word) also fire into $\TPJaction$, and this is how the fact that verb comes after subject is recorded in the synapses between $\Subj$ and $\TPJaction$.

\begin{figure} 
    \centering

 \begin{subfigure}[b]{0.49\textwidth}
    \includegraphics[width=0.99\linewidth,trim={3.5cm 6.5cm 9cm 5cm},clip]{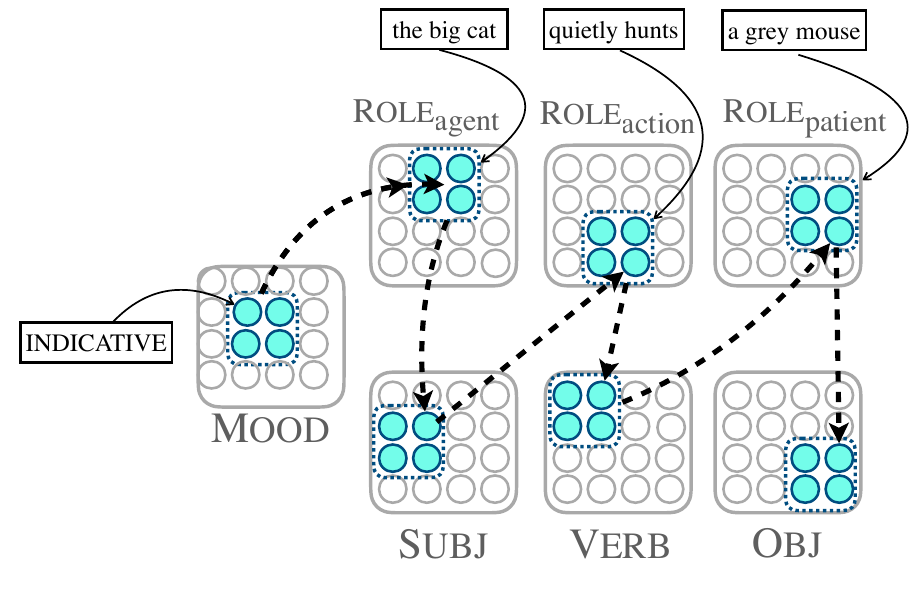}
    \caption{}
    \label{fig:word-order-example}
\end{subfigure}
 \begin{subfigure}[b]{0.49\textwidth}
\includegraphics[width=0.99\linewidth]{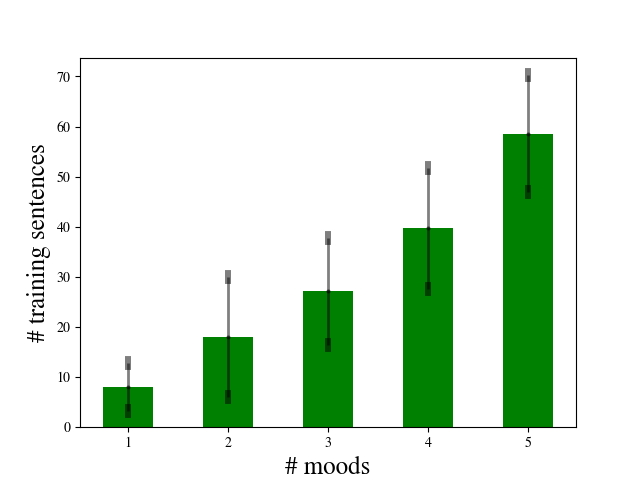}
\caption{}
\label{fig:training-vs-num-moods}
 \end{subfigure}
 \\ 
 \begin{subfigure}[b]{0.9\textwidth}
 \centering
\includegraphics[width=0.6\linewidth]{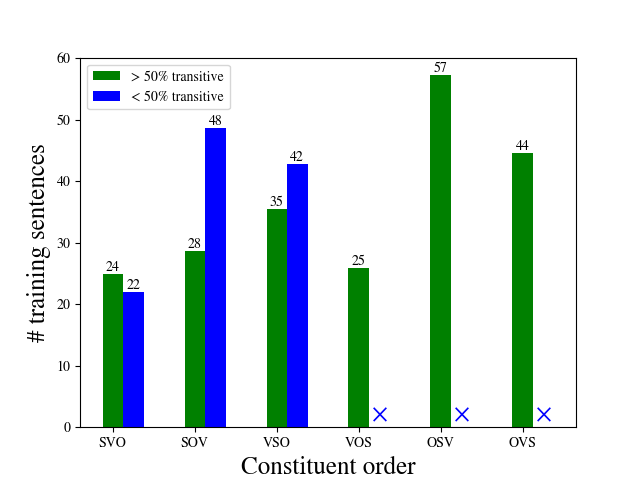}
\caption{}
\label{fig:word-orders}
\end{subfigure}
\caption{(a) An example of the chain of assemblies that fires in a SVO language: the firing of an assembly in $\Mood$ corresponding to the indicative mood activates the $\TPJ$ area with highest input, in this case $\TPJagent$; the assembly in $\TPJagent$ fires, activating an assembly in $\Subj$. When the {trigger assembly} is fired, inhibiting all three \TPJ\ areas, the next one to fire is the one with greatest input from $\Subj$: $\TPJaction$, and so on. In the labels of this example, we imagine an expanded system where the assemblies in each \TPJ\ area can represent whole \emph{phrases,} to suggest how our algorithm for constituent order learning and production is not confined to $3$-word sentences. (b) Our system can learn languages with {multiple} word-orders corresponding to distinct grammatical moods (each mood corresponding to a distinct assembly in the $\Mood$ area). Intuitively, each mood corresponds to a distinct \emph{chain} of assemblies between \TPJ\ and $\Subj, \Verb, \Obj$. We average over $10$ runs per number-of-moods, revealing a roughly linear trend, with somewhat steeper increases in the difficulty to learn the $4$th and the $5$th mood. (c) We compare the number of sentences needed to learn different constituent orders. The algorithm succeeds for any order if more transitive sentences are presented than intransitive sentences, but we find that {significantly more sentences are needed} to learn OSV and OVS (averaging over $20$ runs per order, with $70\%$ transitive input). If the proportion of transitive sentences is less than half, it is easy to show via rigorous proof that the orders VOS, OSV, or OVS {\em cannot} be learned by this system even with unbounded input, and so either an altogether different mechanism, or predominance of transitive input sentences must account for the acquisition of syntax in such languages --- note that these are in fact the three rarest constituent orders among world languages. }
\label{fig:word=order-exp}
\end{figure}

\paragraph{The experiment.} We fix a language and a mood of the language, together determining one of the six possible word orders, and we input a sequence of random transitive and intransitive sentences with this order.
For the experiment to be successful, the proportion of intransitive sentences should not be too high (see the results in Figure \ref{fig:word-orders}); this is to be expected, since an overwhelmingly intransitive input --- and consequently a dearth of the patient constituent --- is obviously detrimental to the learning of the word order.  Every $10$ sentences, we probe by running a generation experiment: we test, upon the presentation of a randomly generated novel scene (e.g., a dog eating a cookie, presented as three assemblies in the three \TPJ\ areas), whether the device will generate the appropriate sentence with the correct word order: ``the dog eats the cookie'' (the details of generation are presented next). Success of the overall learning experiment for this mood and word order is defined as the successful generation of a transitive sentence and another intransitive sentence that are both sampled randomly and withheld during training.

\paragraph{Generation.}  All previously described functions of the \NEMO\ language system were {\em reactive,} in the sense that they were initiated by external stimuli: word assemblies firing and sensory data displayed.  Generation is different; for example,  the human brain has the ability and means to {\em not} generate the sentence ``the dog runs'' even though the dog {\em is} running.  Therefore, generation must involve processes, presumably in the prefrontal cortex, implementing the decision to generate.  We shall model these processes in the simplest possible way: we assume that there is an assembly, in a new area, which, by firing, inhibits all three \TPJ\ role areas for a single step.  We call this assembly the {\em trigger}.

We begin with a scene (imagined or seen) presented in the same way as in training: a lexical assembly for each thematic role has been  projected into the respective \TPJ\ area, with an assembly in $\TPJscene$ synaptically connected to them.  Also, the intended mood is determined by an assembly in the $\Mood$ area.  These two assemblies, in $\Mood$ and $\Scene$, keep firing at every step throughout generation.  Besides that, the generation algorithm is simply the following.

\begin{itemize}
\item[] repeat:
\item[] to generate the next word, fire the trigger assembly
\item[] until the whole sentence has been generated
\end{itemize}



Alternatively, generation could be implemented by firing the trigger only once, with $\Phon$ causing the trigger to fire through synaptic connections after every generated word.  

The simplicity of the algorithm belies quite a bit of complexity. As soon as the \TPJ\ role areas are disinhibited after the single inhibition step, they will receive synaptic input from either (1) mood, (in the case of the first word) or (2) one of the syntactic areas.  The role area with the most synaptic input will be selected, and \emph{this will be the next constituent to fire,} an effect achieved through plasticity during the phase of syntax learning. This constituent activates another assembly in the syntactic areas which has highest synaptic connectivity to the {next} constituent. However, the current constituent will continue to receive the most input from its own recurrent firing, and because the role areas are in mutual inhibition, will continue to fire until the next time all the role areas are inhibited.  This will happen as soon as the next word is about to be generated.


\subsection{Results of the experiment on word order and generation}
We determine how many sentence presentations are needed in order for the model to be able to generate sentences with the correct word order in each mood, for randomly chosen subject, verb, and object, as a function of the size of the lexicon and the number of moods. We withhold {one} particular transitive sentence and {one} intransitive sentence from the training set, and test after each training sentence whether the model generates the withheld sentences in the correct order if we set up the corresponding scene in the \TPJ\ areas and generate. The results of our experiments are shown in Figure \ref{fig:word=order-exp}.  The device succeeds in learning the constituent order, for several languages and moods with different word orders, after the presentation of a number of sentences that appears to grow linearly with the number of moods, and with the size of the lexicon (see Figure \ref{fig:training-vs-num-moods}).

The word orders of thousands of languages in the world have been studied extensively, see, for example, the atlas  \cite{wals}, and it is of interest to revisit this knowledge in light of these results.  One striking observation from the data already mentioned is that of the six possible word orders, the two starting with object --- OVS and OSV --- are by far the two rarest, with VOS coming a close third.  This fact would be easy to explain if the human brain's system for learning word orders were structured as the one presented here: Our experiments show that these three word orders are more difficult to learn, in the sense that many more input sentences are needed ( \ref{fig:word-orders}), and a training corpus that is especially rich in transitive verbs is required (something that is not needed for the other word orders). 
{\em Ergative languages}, in which, in intransitive sentences, the role of object and subject is reversed, do not present our system with extra difficulties.

Finally, there are many languages (about $17$\% of the languages in \cite{wals}) with no fixed word order, in which the order of the constituents of a sentence is determined by considerations of emphasis and style, as well as languages in which a dominant word order exists but is less strict.  We believe that a simple extension of our system can handle such languages through randomization and via the mechanism of additional inputs initiating distinct order chains (as the different \Mood\ assemblies do in the current model).




\subsection{Hierarchy in Language.}  
\begin{figure}
 \begin{subfigure}[b]{0.6\textwidth}
    \centering
    \includegraphics[width=0.5\linewidth]{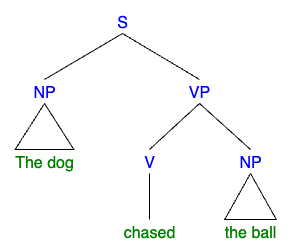}
    \caption{}
    \label{fig:melville}
\end{subfigure}
 \begin{subfigure}[b]{0.4\textwidth}
\includegraphics[width=0.8\linewidth,trim={10cm 8cm 8.5cm 6cm},clip,scale=0.5]{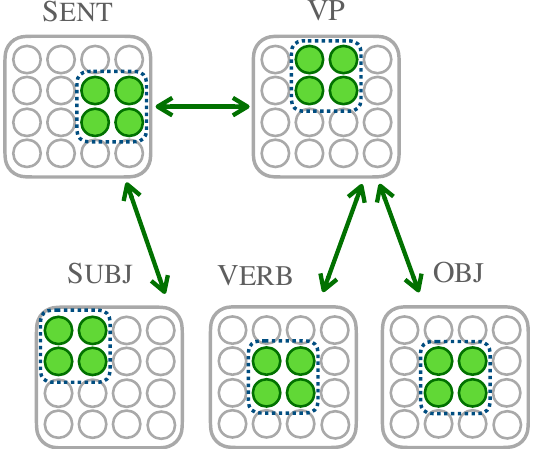}
\caption{}
\label{fig:extra-syntax}
 \end{subfigure} 
 \\
 \begin{subfigure}[b]{0.9\textwidth}
 \centering
\includegraphics[width=0.6\linewidth]{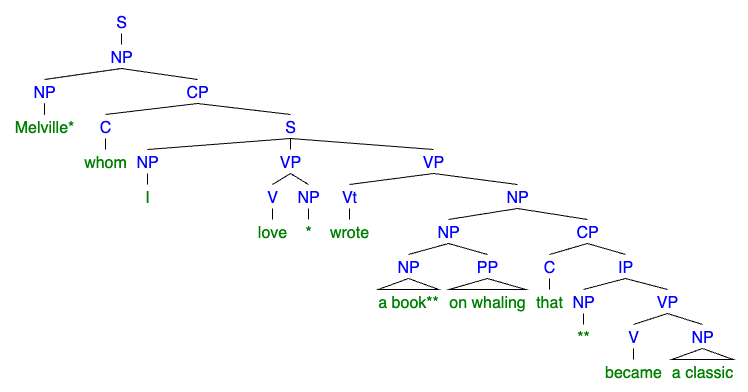}
    \caption{}
\end{subfigure}

 \caption{(a) The syntax tree of a simple sentence, with subject noun phrase, and verb phrase composed of a verb and object noun phrase. (b) An extension of the language acquisition system that can create hierarchical structure (namely, representations of the verb phrase (VP) and of the whole sentence) through the addition of two additional areas (VP and {\sc Sent}, representing subareas of Broca's area. Fibers are shown as green arrows. Firing of assemblies in {\sc Verb} and {\sc Obj} into VP activates an assembly representing the verb phrase; similarly, firing from VP and {\sc Subj} into {\sc Sent} results in a firing assembly representing the whole sentence (or the ``root'' of the syntactic tree at S in subfigure (a)).  (c) The syntax tree of a sentence, showing a hierarchy of embedded structure. }
 \label{fig:hierarchical}
\end{figure}

So far, we have treated language acquisition as learning how to comprehend and generate sentences, that is, short {\em sequences} of words. However, it is generally believed that language also has a dual {\em hierarchical} nature.  At the most basic level, each sentence has a {\em syntax tree} that reveals its syntactic structure (see Figure \ref{fig:hierarchical}). Furthermore, sentences can be embedded recursively in one another, resulting in larger hierarchies.  For example, the utterance {\em ``Melville, whom I love, wrote a book on whaling that became a classic''} is the nested combination of three individual sentences (see Figure \ref{fig:melville}). The relative importance of the hierarchical and sequential aspects of language is the subject of a well-known discord in linguistics; see, for example, \cite{Poeppel}, as well as the responses to that paper, for a recent recurrence.  

Can our implementation of language acquisition shed light on the role of hierarchies in language? To quote noted neurolinguist Angela Friederici \cite{FriedericiChomskyBook}, {\em ``It would really be crucial to show that there is a limit for ordered sequences and that beyond that limit a (hierarchical) nesting is necessary.''}.  Interestingly, in recent work on the implementation in \NEMO\ of planning \cite{Planning} and sequence memorization \cite{DPV25} it has been observed that, precisely as Friederici has suggested, it is difficult to maintain sequences of assemblies longer than a limit.  The limit varies with the parameters of the \NEMO\ model, and ranges between 20 and 40. This difficulty suggests that a full language system implemented in \NEMO\ { must} also deal with the hierarchical aspects of language.

The language acquisition system presented so far can be extended in a very natural way to support the creation of hierarchical structures. Much of the system in Figure \ref{fig:model-full} --- that is, the areas for phonology, verbs, nouns, moods, and roles, omitting the semantic areas --- models parts of the human language system that are thought to be located in the left medial temporal lobe (lMTL) of the brain, in what is broadly referred to as {\em Wernicke's area}.  Now we can postulate that the three syntactic areas ({\sc Subj}, {\sc Verb}, {\sc Obj}) are located in {\em Broca's area} in the left frontal gyrus; recall that parts of Broca's area are thought to be involved in syntactic processing and generation \cite{Friederici}, and the two areas of Broca and Wernicke are connected by the arcuate fasciculus (AF) axon bundle.  That is, we postulate that the nine fibers connecting the role areas and the syntactic areas in Figure \ref{fig:model-full}, implementing word order, are parts of the AF.  In addition to the three areas {\sc Subj}, {\sc Verb}, and {\sc Obj}, we can assume two more areas named {\sc Sent} and {\sc VP}, also meant to be located in Broca's area, standing for {\em sentence} and {\em verb phrase,} respectively. These five areas are connected through fibers as follows: {\sc Verb} and {\sc Obj} are both connected to VP, while {\sc Subj} and VP are both connected to {\sc Sent} (see Figure \ref{fig:extra-syntax}).  This structure mirrors the skeleton of the basic syntax tree of a transitive sentence, and the same structure can also work for intransitive sentences.  The effect of this new arrangement is that, during both comprehension and production of a sentence, the basic syntax tree of the sentence is created through synaptic connections between the assemblies in these areas. As for embedded structure, like the relative clauses of the sentence in Figure \ref{fig:melville}, previous work on the theory of parsing in \NEMO\ showed that such a structure can be handled with an additional area (called {\sc Dep} in previous work) whose assemblies link a word to its dependent clause \cite{CenterEmbedding}. However, parsing and generating such hierarchically structured sentences presents novel challenges, which need further work. 

\section{Discussion}\label{sec:hypothesis}
We presented what we believe is the first biologically plausible system that can learn the rudiments of natural language --- lexicon as well as basic syntax and semantics --- through the presentation of a modest amount of grounded sentences in any language.  Of course, many facets of language are not covered by our system: The learning of speech sounds and phonology has been bypassed, while adjectives, adverbs, and functional categories are not considered. We believe that many of these missing aspects can be accommodated by simple extensions of the present system. {\em Abstract words} --- words that are not directly grounded in sensorimotor experience, such as ``bravery,'' ``available,'' and ``negotiate'' --- are an important next step that will require, among others, the development of a representation scheme for such words and the introduction of new semantic brain areas.  There are many current cognitive linguistic theories on how abstract words are dealt with in the human brain, see, for example, \cite{Borghi2017}, which can guide progress on this front.  Other sophisticated aspects of linguistic competence, such as resolution of ambiguity, grammaticality judgment, metaphor, and next-word prediction, to mention a few, are very interesting further challenges, and we are confident  that they can be addressed in our framework, even though some of them may require certain simple and biologically plausible extensions of the \NEMO\ model (recall the discussion of moods in the previous section). 

Further research goals would be learning the social uses of language, as well as its interaction with aspects of social cognition such as agency and the theory of mind, which presuppose new kinds of representations and mechanisms.  In addition, it would be very intriguing to experiment with interactions between \NEMO\ artifacts and large language models (LLMs) as a means to help the former grow in these more demanding directions.  Beyond language, {\em reasoning} in its various genres --- logical, heuristic, categorical, probabilistic, etc.~~-- is an important next challenge for \NEMO\ artifacts.

Our system can be extended to handle {\em multilinguality:} A new area can be added containing an assembly for each language known to the speaker, very much like the representations of grammatical moods in the $\Mood$ area.  This assembly would be active while learning, or using, the corresponding language.  Initial experimentation with this idea is encouraging, but more work is needed.

We believe that our language acquisition system is biologically plausible not only because it is strictly based on the \NEMO\ framework, but also because its {\em tabula rasa} brain architecture --- the areas and the fibers connecting them --- can be plausibly delivered by mammalian developmental processes.  One possible exception are the assemblies in the $\Mood$ area, each representing a different mood of the language. It can be argued that such a mood assembly can be created during learning on the basis of various sensory and mental-state cues associated with the mood (e.g., the tension in the discourse and the special tone of voice when a question is asked).  However, more work is needed to fully justify this point.

Notwithstanding  the important and challenging research directions that lie ahead and the limitations mentioned above, the artificial system for basic language acquisition based on \NEMO\ we present here is the first of its kind, and, besides its value as a proof of concept, may be of use to neurolinguistic research.  Its design decisions are intended to largely reflect current neurolinguistic thinking, and hence the resulting artifact is a concrete hypothesis that can possibly help guide further research on the human language system through the generation of testable predictions and the deployment of more advanced versions.

\clearpage

\def\SubjNP{\Subj_{\text{NP}}}
\def\ObjNP{\Obj_{\text{NP}}}
\def\VP{\textsc{VP}}

\newpage

\bibliography{neurips2023}
\bibliographystyle{acl_natbib}

\newpage

\end{document}